\documentclass[letterpaper, 10 pt, conference]{ieeeconf}  

\IEEEoverridecommandlockouts                              

\usepackage{amsmath} 
\usepackage{amssymb}  
\usepackage{siunitx}
\usepackage{tikz}
\usetikzlibrary{fit}
\usetikzlibrary{positioning,arrows}
\usetikzlibrary{arrows.meta}
\usetikzlibrary{shapes.misc}
\usepackage{wasysym}
\usepackage{graphicx}
\usepackage{varwidth}
\usepackage{subcaption}
\usepackage{cite}
\usepackage{hyperref}
\usepackage{pbox}
\usepackage{quotes}

\usepackage[font=small]{caption}

\tikzset{
  state/.style={circle,draw,minimum size=6ex},
  arrow/.style={-latex, shorten >=1ex, shorten <=1ex}}

\tikzset{cross/.style={cross out, draw=black, minimum size=2*(#1-\pgflinewidth), inner sep=0pt, outer sep=0pt},
cross/.default={4pt}}

\DeclareMathOperator{\atantwo}{atan2}

\title{\LARGE \bf
Model-based Hand Pose Estimation for Generalized Hand Shape with Appearance Normalization
}

\author{Jan W\"ohlke*$^{1}$, Shile Li*$^{1}$,  and Dongheui Lee$^{1,2}$
\thanks{*Co-first authors}
\thanks{$^{1}$Human-centered Assistive Robotics, Technical University of Munich, Munich, Germany, {\tt\small jan.woehlke@tum.de, li.shile@mytum.de, dhlee@tum.de }}%
\thanks{$^{2}$Institute of Robotics and Mechatronics, German Aerospace Center, Wessling, Germany}%
}

\begin{document}

\maketitle

\begin{abstract}


Since the emergence of large annotated datasets, state-of-the-art hand pose estimation methods have been mostly based on discriminative learning.
Recently, a hybrid approach has embedded a kinematic layer into the deep learning structure in such a way that the pose estimates obey the physical constraints of human hand kinematics.
However, the existing approach relies on a single person's hand shape parameters, which are fixed constants. Therefore, the existing hybrid method has problems to generalize to new, unseen hands.
In this work, we extend the kinematic layer to make the hand shape parameters learnable. In this way, the learnt network can generalize towards arbitrary hand shapes.
Furthermore, inspired by the idea of Spatial Transformer Networks, we apply a cascade of appearance normalization networks to decrease the variance in the input data. The input images are shifted, rotated, and globally scaled to a similar appearance. 
The effectiveness and limitations of our proposed approach are extensively evaluated on the Hands 2017 challenge dataset and the NYU dataset.

\end{abstract}

\section{INTRODUCTION}

Hand pose estimation is an important requirement for many tasks in fields such as human computer interaction or augmented reality. Therefore, the 3D hand pose estimation problem has attracted many researchers' interest in the last ten years\cite{Erol2007}\cite{Supancic2015}\cite{Yuan2017c}.
The availability of cheap commercial depth cameras has especially increased the interest. 
However, it still remains challenging due to several reasons: the kinematic complexity of the hand, which results in a large number of DoFs, self-occlusions, different viewpoints, and shape variations across different persons.

Hand pose estimation approaches can be divided into three categories: 1) the generative, model-driven approaches that fit a hand model to the image observations by minimizing a cost function\cite{Khamis2015}\cite{Sharp2015}\cite{Sridhar2015}\cite{Tan2016}, 2) the discriminative, data-driven approaches that directly predict the 3D joint locations from the images\cite{Chen2017b}\cite{Deng2017}\cite{Ge2016}\cite{Ge2017}\cite{Guo2017b}\cite{Madadi2017a}\cite{Oberweger2015a}\cite{Oberweger2017}\cite{Moon2017}, and 3) the hybrid approaches that combine discriminative and generative elements\cite{Oberweger2015b}\cite{Tompson2014}\cite{Wan2017a}\cite{Zhou2016}. 

Discriminative methods play an important role because they are needed to initialize generative tracking methods and to recover in the case of tracking failure. State-of-the-art discriminative methods use deep learning components such as 2D \cite{Chen2017b}\cite{Ge2016}\cite{Guo2017b}\cite{Madadi2017a}\cite{Oberweger2015a}\cite{Oberweger2015b}\cite{Oberweger2017}\cite{Tompson2014}\cite{Ye2016}\cite{Ye2017}\cite{Zhou2016} or 3D \cite{Deng2017}\cite{Ge2017}\cite{Moon2017} Convolutional Neural Networks (CNN) that might incorporate residual modules \cite{Chen2017b}\cite{Guo2017b}\cite{Moon2017}\cite{Oberweger2017}. Relying on a large annotated training dataset, the discriminative methods either directly regress joint locations \cite{Chen2017b}\cite{Deng2017}\cite{Ge2017}\cite{Madadi2017a}\cite{Oberweger2015a}\cite{Oberweger2015b}\cite{Oberweger2017}\cite{Ye2016}\cite{Ye2017}\cite{Zhou2016} or output a probability density map for each joint \cite{Ge2016}\cite{Moon2017}\cite{Tompson2014}.

\begin{figure}[t]
\vspace{-0.2cm}
\centering
\includegraphics[width=0.35\textwidth]{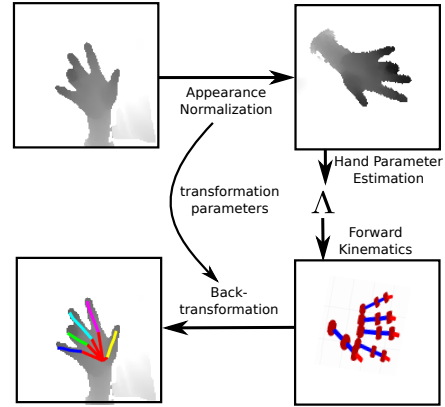}

\caption[Overview] { \small Overview of our approach. The input images are appearance-normalized. Then, hand parameters $\boldsymbol{\Lambda}$ are estimated. These are fed into a kinematic layer that maps them to joint locations which are back-transformed into the initial coordinate system.}

\label{fig:overview}
\vspace{-0.8cm}
\end{figure}

Most discriminative approaches do not explicitly consider the kinematics and physical motion constraints of the hand. As a result, kinematically implausible hand pose estimates can occur. For example, the physical limits of the finger joint angles can be violated. To ensure physically plausible estimates, hybrid models that incorporate a generative component can be used. For example, Zhou et al. \cite{Zhou2016} integrate a generative forward kinematics hand model into their deep learning approach and impose physical constraint losses on the estimated hand parameters. However, in \cite{Zhou2016}, the palm shape and bone lengths are fixed to a specific user so that the approach cannot generalize towards new hand shapes.

In this work, we extend \cite{Zhou2016} by proposing two novel elements, so that it can generalize to arbitrary hand shapes. Our main contributions are (see Fig.~\ref{fig:overview} for an overview):
\begin{itemize}
\item A novel kinematic layer, where the hand shape parameters (palm shape, bone lengths) become variables whose values are regressed from the input image.
\item A cascade of networks for appearance normalization to subsequently re-center, rotate, and re-scale the input hand images so that they approximately have a normalized appearance.
\item To evaluate the effectiveness and limitation of the proposed ideas, we extensively test the different network components standalone as well as in combination on the Hands 2017 challenge \cite{Yuan2017b} and the NYU \cite{Tompson2014} datasets.
\end{itemize}

\section{RELATED WORK}

In this section, we first review previous works to explicitly consider the hand geometry including the kinematic structure of the hand as well as the physical constraints in hand pose estimation. Furthermore, some works on learning spatial transformations with neural networks are reviewed.

\subsection{Considering Hand Geometry in Hand Pose Estimation}

There are three main strategies to explicitly consider the hand geometry in deep-learning-based hand pose estimation: 1) a hand model component is added to the network architecture, 2) the joint location estimates are post-processed, or 3) constraints are added to the network loss formulation.

1) One strategy is to directly enforce the structure of the hand within the network architecture \cite{Oberweger2015a}\cite{Oberweger2017}\cite{Wan2017a}\cite{Zhou2016}. The DeepPrior(++) approach \cite{Oberweger2015a}\cite{Oberweger2017} integrates a pose prior into the network. Instead of directly predicting 3D joint locations, the Convolutional Neural Network (CNN) predicts hand pose parameters in a lower dimensional space. The reconstruction to the original pose space is a linear embedding whose weights are initialized with the major components of a Principal Component Analysis (PCA) of the hand poses in the training data. However, a linear embedding can only approximate the highly non-linear hand kinematics. Therefore, the DeepModel architecture presented in \cite{Zhou2016} integrates a generative non-linear hand model layer into the CNN, instead. The hand model layer implements forward kinematics, taking hand parameters predicted by the CNN as input and mapping them to 3D joint locations. Physical constraint losses on the estimated hand parameters ensure the physical validity of estimated poses. However, the bone lengths and palm shape of the kinematic model are fixed to a single user. The network cannot generalize to new hand shapes. While \cite{Zhou2016} uses a parametric hand model, the Crossing Nets architecture \cite{Wan2017a} learns the hand model itself by combining two generative neural networks with a shared latent space in a multi-task learning setting. The combined networks model the generation process of 3D hand poses and depth images, respectively, as well as their statistical relationship. As a result, the combined model is capable of generating depth images given a hand pose as well as inferring a hand pose from a given depth image.

2) Another strategy is to enforce the hand geometry by appropriate post-processing of estimates \cite{Tompson2014}\cite{Ye2016}. In \cite{Tompson2014}, the output joint location heat maps are refined to 3D joint locations by optimizing a hand model using Particle Swarm Optimization (PSO). However, such post-processing that is separated from the training procedure is sub-optimal. Ye et al. \cite{Ye2016} minimize an energy function using PSO for enforcing kinematic constraints between the kinematic hierarchy levels of their hierarchical hand pose estimation approach. 

3) As a third strategy, Sun et al. \cite{Sun2017} integrate the kinematic constraints into the loss of their human pose regression network. They use a bone-based instead of a joint-based representation of the pose. The pose is structured as a tree with a root and bones that encode the 3D distances between two subsequent joints in the kinematic tree. A compositional loss formulation is proposed that exploits the geometric joint connection structure and encodes long range interactions in the pose, as an error in one bone propagates along the kinematic chain. Each bone is constrained by multiple paths. When considering the distances between arbitrary joints, the pose structure can be fully exploited.

In our approach, the physical validity of the hand pose estimates is enforced implicitly by the kinematic hand model layer as well as explicitly by physical constraint losses on the finger joint angles. No post-processing is needed. Furthermore, the variation across different hand shapes is considered by making the hand model parameters learnable.

\subsection{Learning Spatial Transformations}

The idea of using neural networks to learn spatial transformations that are applied to the input images of other neural networks, in order to make them invariant to translation, rotation, scaling, or more generic warping, is not entirely new. Hinton et al. \cite{hinton2011transforming} propose a transforming autoencoder as a generative model that models 2D affine transformations. The generative model learns to generate a transformed image of the input image, where the target pose is defined during the training. Zimmermann et al. \cite{zimmermann2017} estimate the hand pose in a normalized coordinate system. The global transformation parameters are regressed separately. Jaderberg et al. \cite{Jaderberg2015} introduce a dynamic mechanism, the Spatial Transformer, that is trained end-to-end with the rest of the network without changing the loss function. A localization network regresses the transformation parameters from the input image, which are then used by the grid generator to transform a regular grid into a sampling grid. This sampling grid is applied to the input image to obtain the warped output image.

On the contrary, our approach applies a cascade of networks to regress the transformation parameters. The transformation parameters are tailored to hand pose estimation. Furthermore, the later cascade stages benefit from the normalization of the earlier stages. Besides that, the individual networks are trained with ground truth transformation parameters, while earlier cascade stages are fixed. The cascade is not further trained but used for inference when training the actual hand pose estimation network on top of it.

For our approach, we extend the idea of \cite{Oberweger2017} to use a CNN to predict the deviation between the hand center of mass and the middle finger metacarpophalangeal (MCP) joint, to re-center the input images, as well as the idea of using a CNN to predict an overall scaling factor of the hand, to re-scale it to unit scale, that was used in our groups' Hands 2017 challenge submission \cite{Yuan2017c}. Furthermore, the RotNet for predicting the rotation of the middle finger is newly introduced.

\section{DATASET PROCESSING}

\subsection{Offline Pre-processing}

First, a coarse 3D bounding box containing the hand is determined from the joint location ground truth annotation $\boldsymbol{J}^{\text{gt}}$ (or provided for the Hands 2017 challenge test set). The corner points are calculated from the minimum and maximum pixel coordinates across all hand joints. Clustering the depth values inside the bounding box gives the depth center of mass. The bounding box center defines the planar hand center of mass (COM). Afterwards, a fixed-size 3D cube centered on the COM is extracted from the depth map. It includes margins for a later re-centering of the images.

Before re-projecting the 3D cube to a depth image, the 3D cube is rotated so that the hand center of mass (COM) projects to the center of the image plane. This normalizes the hand image appearance. The needed rotation matrix $\boldsymbol{R}_\text{CAM}$ is determined based on the COM coordinates:
\begin{eqnarray}
\alpha_y &=& \atantwo\left(\text{COM}_x,\text{COM}_z\right)\\
\widetilde{\text{COM}} &=& \boldsymbol{R}_y\left(-\alpha_y\right)\cdot\text{COM}\\
\alpha_x &=& \atantwo\left(\widetilde{\text{COM}}_y,\widetilde{\text{COM}}_z\right)\\
\boldsymbol{R}_\text{CAM} &=& \boldsymbol{R}_y\left(-\alpha_y\right)\cdot\boldsymbol{R}_x\left(\alpha_x\right)
\label{eq:rot}
\end{eqnarray}
All 3D points within the extracted 3D depth cube as well as the ground truth joint locations $\boldsymbol{J}^{\text{gt}}$ are rotated by $\boldsymbol{R}_\text{CAM}$. Furthermore, the COM depth value is subtracted from the ground truth joint locations $\boldsymbol{J}^{\text{gt}}$ so that their representation is relative to the COM.

Finally, the 3D depth cube representation is back-projected to a depth image and re-sized to $176\times 176$ pixels. For smoothing, $3\times 3$ median filtering is applied. Afterwards, the depth values are normalized to the range $\lbrack -1,1\rbrack$.

\subsection{Online Augmentation}

During training, online data augmentation is applied. The images are scaled, rotated, and translated by random factors drawn from the following distributions:
\begin{itemize}
\item {\bf scaling:} normal $\mathcal{N}\left(1.0,0.075\right)$ within range $\lbrack 0.75,1.25 \rbrack$
\item {\bf rotation:} uniform $\mathcal{U}\left(-\SI{180}{\degree},\SI{180}{\degree}\right)$ 
\item {\bf translation:} normal $\mathcal{N}\left(0,\SI{4}{\milli\metre}\right)$ within range $\lbrack -15,15\rbrack$ \si{\milli\metre} for $x$-, $y$-, and $z$-direction individually
\end{itemize}

\section{HAND POSE ESTIMATION APPROACH}

Figure~\ref{fig:cascade} depicts our hand pose estimation approach. In the following, its main components are described in detail.

\subsection{Appearance Normalization Pipeline}
\label{sec:AppNorm}

In order to increase the robustness of hand pose estimation towards different appearances of the hand images, a cascade of three neural networks and suitable image processing, the appearance normalization pipeline, is introduced (see Fig.~\ref{fig:cascade}). It approximately normalizes the hand images with respect to translations, in-plane rotations, and overall scaling.

First, the $176\times 176$ input depth images are passed through the BoxNet, which estimates the 3D translation $\boldsymbol{t}$ of the middle finger MCP joint with respect to the COM (initial image center). Based on the estimate $\boldsymbol{t}^\text{est}$, a $128\times 128$ image centered on the estimated middle finger MCP joint is cropped (see Fig.~\ref{fig:recrop}). The normalized depth values are re-centered on the middle finger MCP joint depth.
\setcounter{figure}{2}
\begin{figure}[htb]
\begin{subfigure}[htb]{0.24\textwidth}
\centering
\begin{tikzpicture}

\draw (-1.76,1.76) -- (1.76,1.76) -- (1.76,-1.76) -- (-1.76,-1.76) -- cycle;

\draw[red] (-1.06,0.86) -- (1.5,0.86) -- (1.5,-1.7) -- (-1.06,-1.7) -- cycle;

\draw[black,fill=black] (0,0) circle (.5ex);
\draw[red,fill=red] (0.22,-0.42) circle (.5ex);

\node at (0,0.25) {COM};
\node at (0.72,-0.42) {\textcolor{red}{MCP}};

\node at (-0.1,-0.25) {\textcolor{blue}{$\boldsymbol{t}$}};

\node at (-1.06,1.46) {\small 176x176};
\node at (-0.36,0.56) {\small \textcolor{red}{128x128}};

\draw [arrows={-Triangle[length=0.1cm, width=0.1cm]},blue] (0,0) -- (0.22*0.9,-0.42*0.9);

\end{tikzpicture}
\caption{Re-centering}
\label{fig:recrop}
\end{subfigure}
\begin{subfigure}[htb]{0.24\textwidth}
\centering
\begin{tikzpicture}

\draw [arrows={-Triangle[length=0.3cm, width=0.3cm]}] (0,0) -- (2.6,0);
\draw [arrows={-Triangle[length=0.3cm, width=0.3cm]}] (0,0) -- (0,-2.6);

\draw [arrow, bend angle=30, bend left]  (1.25,0.125) to (1,-0.8);

\draw[red] (0,0) -- (1.875,-1.25);

\node at (2.6,0.25) {$x$};
\node at (0.25,-2.6) {$y$};
\node at (0,0.25) {MCP};
\node at (2,-1) {TIP};
\node at (0.75,-0.25) {$\alpha_z$};

\draw[black,fill=black] (0,0) circle (.5ex);
\draw[black,fill=black] (1.875,-1.25) circle (.5ex);

\end{tikzpicture}
\caption{Rotation}
\label{fig:rotang}
\end{subfigure}
\caption{\small Image re-centering and rotation: {\bf (a)} Re-centered image (red) cropped from input image (black). Translation vector $\boldsymbol{t}$ (in $x$-$y$-plane) shown in blue. {\bf (b)} Rotation angle $\alpha_z$ of middle finger (red). MCP and TIP joint defined in Fig.~\ref{fig:handdh}.}
\vspace{-0.7cm}
\end{figure}
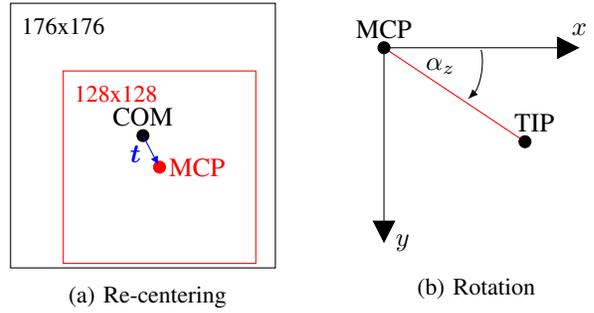

The re-centered images are then passed through the RotNet, which estimates the in-plane $z$-rotation angle $\alpha_z$ of the middle finger with respect to the $x$-axis (see Fig.~\ref{fig:rotang}). Afterwards, the images are rotated around the $z$-axis by the negative estimates $-\alpha_z^\text{est}$ so that the middle finger approximately coincides with the $x$-axis.

Subsequently, the images are passed through the ScaleNet, which estimates an overall scaling factor $s$ of the hand  that is defined as the sum of all 20 hand bone lengths divided by the sum of the average bone lengths across the training set. Based on the estimate $s^{\text{est}}$, the images are re-scaled in all three spatial directions (maintaining normalized size) so that the hand in the image has approximately unit size.

Finally, the images are passed through the PoseNet, which estimates the 3D joint locations $\tilde{\boldsymbol{J}}$ of the approximately appearance-normalized images. It consists of a convolutional or residual network, termed ParamNet (see Sec.~\ref{sec:NetArc}), that estimates hand parameters $\boldsymbol{\Lambda}$, and a kinematic hand model layer (\textit{FKINE}) that implements forward kinematics from hand parameters $\boldsymbol{\Lambda}$ to joint locations $\tilde{\boldsymbol{J}}$ (see Sec.~\ref{sec:KinLayer}).

The joint location estimates $\tilde{\boldsymbol{J}}$ are back-transformed to $\boldsymbol{J}$ by scaling, rotating, and translating the individual joint locations in order to match the original (pre-processed) input images:
\begin{equation}
\boldsymbol{j}_{i,k} = \left(\boldsymbol{R}_z\left(\alpha_{z}^\text{est}\right)\cdot\left(s^{\text{est}}\cdot\tilde{\boldsymbol{j}}_{i,k}\right)\right)+\boldsymbol{t}^{\text{est}}
\end{equation}
The index $i\in \lbrace\text{T},\text{I},\text{M},\text{R},\text{P}\rbrace$ indicates the respective finger, whereas $k\in \lbrace \text{WRIST},\text{MCP},\text{PIP},\text{DIP},\text{TIP} \rbrace$ indicates the finger joint type (see Fig.~\ref{fig:hand}).

Due to our offline pre-processing, the joint location estimates $\boldsymbol{j}_{i,k}$ must be post-processed offline. The stored COM depth is added to the $\boldsymbol{j}_{i,k}$, before they are rotated into the original coordinate system using the rotation matrix $\boldsymbol{R}_\text{CAM}^{-1}$.

Box-, Rot-, and ScaleNet are applied in the inverse order of the corresponding augmentation components (scaling, rotation, translation). Therefore, the appearance normalization pipeline can be interpreted as an approximate inverse function to the augmentation process, which, furthermore, takes out some of the variance in scaling, rotation, and translation that already exists in the un-augmented images.

\setcounter{figure}{1}
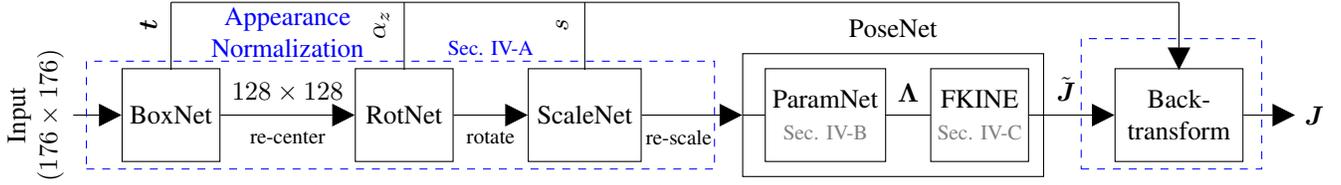
\begin{figure*}
\centering
\begin{tikzpicture}

\coordinate (st0) at (-1,0.7);
\coordinate (en0) at (-1,-0.7);
\node[rectangle,minimum width=0.5cm] [fit = (st0) (en0)] (bx0) {};
\node[align=center,rotate=90] at (bx0.center) {Input\\ $\left(176\times 176\right)$};

\coordinate (st1) at (0.8,0.5);
\coordinate (en1) at (0.8,-0.5);
\node[rectangle,draw,minimum width=1.3cm] [fit = (st1) (en1)] (bx1) {};
\node[align=center] at (bx1.center) {BoxNet};

\coordinate (st2) at (3.9,0.5);
\coordinate (en2) at (3.9,-0.5);
\node[rectangle,draw,minimum width=1.3cm] [fit = (st2) (en2)] (bx2) {};
\node[align=center] at (bx2.center) {RotNet};

\coordinate (st3) at (6.3,0.5);
\coordinate (en3) at (6.3,-0.5);
\node[rectangle,draw,minimum width=1.5cm] [fit = (st3) (en3)] (bx3) {};
\node[align=center] at (bx3.center) {ScaleNet};

\coordinate (st4) at (10.4,0.7);
\coordinate (en4) at (10.4,-0.7);
\node[rectangle,draw,minimum width=4cm] [fit = (st4) (en4)] (bx4) {};
\node[align=center] at (bx4.center) {};
\node at (10.4,1.15) {PoseNet};

\coordinate (st2a) at (9.5,0.5);
\coordinate (en2a) at (9.5,-0.5);
\node[rectangle,draw,minimum width=1.6cm] [fit = (st2a) (en2a)] (bx2a) {};
\node[align=center] at (bx2a.center) {};
\node[align=center] at (9.5,0) {ParamNet\\ {\footnotesize \color{gray} Sec.~\ref{sec:NetArc}}};

\coordinate (st2c) at (11.55,0.5);
\coordinate (en2c) at (11.55,-0.5);
\node[rectangle,draw,minimum width=1.3cm] [fit = (st2c) (en2c)] (bx2c) {};
\node[align=center] at (bx2c.center) {};
\node[align=center] at (11.55,0) {FKINE\\ {\footnotesize \color{gray} Sec.~\ref{sec:KinLayer}}};

\draw (8.4,0) -- (8.7,0);
\draw (10.3,0) -- (10.9,0);
\draw (12.2,0) -- (12.4,0);

\node at (10.6,0.3) {$\boldsymbol{\Lambda}$};

\coordinate (st5) at (14.2,0.5);
\coordinate (en5) at (14.2,-0.5);
\node[rectangle,draw,minimum width=1.7cm] [fit = (st5) (en5)] (bx5) {};
\node[align=center] at (bx5.center) {Back-\\ transform};

\coordinate (st6) at (16,0.7);
\coordinate (en6) at (16,-0.7);
\node[rectangle,minimum width=0.5cm] [fit = (st6) (en6)] (bx6) {};
\node[align=center] at (bx6.center) {$\boldsymbol{J}$};

\node at (2.35,0.3) {$128\times 128$};
\node at (2.35,-0.3) {\footnotesize re-center};
\node at (5.05,-0.3) {\footnotesize rotate};
\node at (7.55,-0.3) {\footnotesize re-scale};
\node at (12.7,0.35) {$\tilde{\boldsymbol{J}}$};

\node[align=center,rotate=90] at (0.5,1.2) {$\boldsymbol{t}$};
\node[align=center,rotate=90] at (3.6,1.2) {$\alpha_z$};
\node[align=center,rotate=90] at (6,1.2) {$s$};

\draw [arrows={-Triangle[length=0.3cm, width=0.3cm]}] (-0.5,0) -- (bx1);
\draw [arrows={-Triangle[length=0.3cm, width=0.3cm]}] (bx1) -- (bx2);
\draw [arrows={-Triangle[length=0.3cm, width=0.3cm]}] (bx2) -- (bx3);
\draw [arrows={-Triangle[length=0.3cm, width=0.3cm]}] (bx3) -- (bx4);
\draw [arrows={-Triangle[length=0.3cm, width=0.3cm]}] (bx4) -- (bx5);
\draw [arrows={-Triangle[length=0.3cm, width=0.3cm]}] (bx5) -- (bx6);
\draw [arrows={-Triangle[length=0.3cm, width=0.3cm]}] (14.2,1.5) -- (14.2,0.6);

\draw (0.8,1.5) -- (14.2,1.5);
\draw (0.8,0.6) -- (0.8,1.5);
\draw (3.9,0.6) -- (3.9,1.5);
\draw (6.3,0.6) -- (6.3,1.5);

\coordinate (st7) at (3.85,0.6);
\coordinate (en7) at (3.85,-0.6);
\node[rectangle,draw,dashed,blue,minimum width=8.35cm] [fit = (st7) (en7)] (bx7) {};

\coordinate (st8) at (14.1,0.9);
\coordinate (en8) at (14.1,-0.6);
\node[rectangle,draw,dashed,blue,minimum width=2.4cm] [fit = (st8) (en8)] (bx8) {};

\node[align=center,blue] at (2.35,1.1) {Appearance\\ Normalization};
\node[align=center,blue] at (5.05,0.9) {\footnotesize Sec.~\ref{sec:AppNorm}};

\end{tikzpicture}
\vspace{-0.7cm}
\caption{Our model-based hand pose estimation approach with appearance normalization.}
\label{fig:cascade}
\vspace{-0.7cm}
\end{figure*}

Four training runs are necessary to train the full pipeline including PoseNet. The four networks are trained one after the other on top of each other. The parameters of the networks in front of the currently trained one are fixed. Box-, Rot-, and ScaleNet are trained with ground truth values $\boldsymbol{t}^{\text{gt}}$, $\alpha_z^{\text{gt}}$, and $s^{\text{gt}}$ calculated from the joint location ground truth.  

\subsection{Network architectures}
\label{sec:NetArc}

Due to its simplicity and computational efficiency, we use a basic CNN architecture (see Fig.~\ref{fig:cnn}) for Box-, Rot-, and ScaleNet with $l=3$, $l=1$, and $l=1$ output units, respectively. Furthermore, this CNN architecture is used for ParamNet in Sec.~\ref{sec:eval_1} and Sec.~\ref{sec:eval_2}. For a higher estimation accuracy, we employ a residual network \cite{He2015b}\cite{He2016} as ParamNet in Sec.~\ref{sec:eval_3}. The architecture is based on the ResNet50 architecture and modified to apply it to hand pose estimation (see Fig.~\ref{fig:resnet}).

All network layers use ReLU units whose weights are initialized from the uniform distribution $\mathcal{U}\left(-\frac{2}{\sqrt{m}},\frac{2}{\sqrt{m}}\right)$ ($m$: the number of layer inputs) in case of the CNN  and from a normal distribution with standard deviation chosen according to \cite{He2015a} in case of the ResNet.
\setcounter{figure}{3}
\begin{figure}[!h]
\vspace{-0.3cm}
\centering
\begin{subfigure}{0.49\textwidth}
\centering
\includegraphics[scale=0.7]{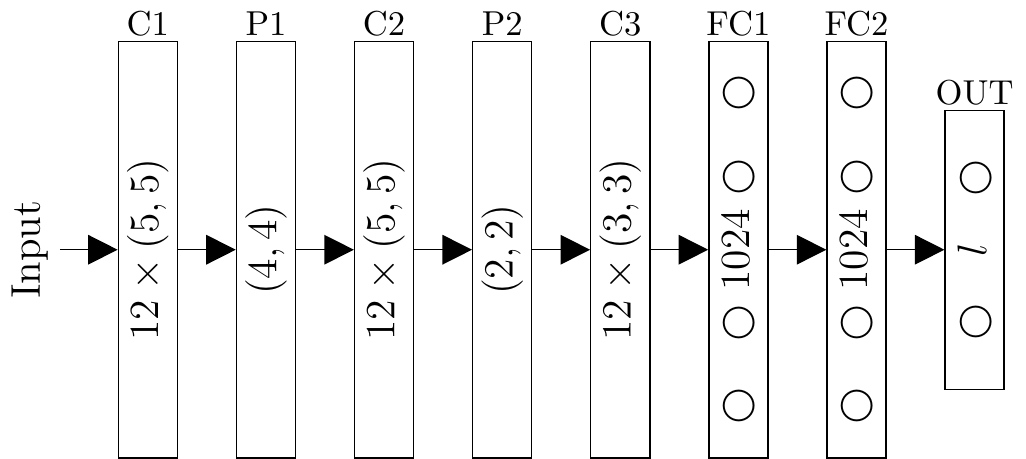}
\caption{Basic CNN architecture used for Box-, Rot-, and ScaleNet as well as ParamNet in Sec.~\ref{sec:eval_1} and Sec.~\ref{sec:eval_2}}
\label{fig:cnn}
\end{subfigure}

\begin{subfigure}[htb]{0.49\textwidth}
\centering
\includegraphics[scale=0.7]{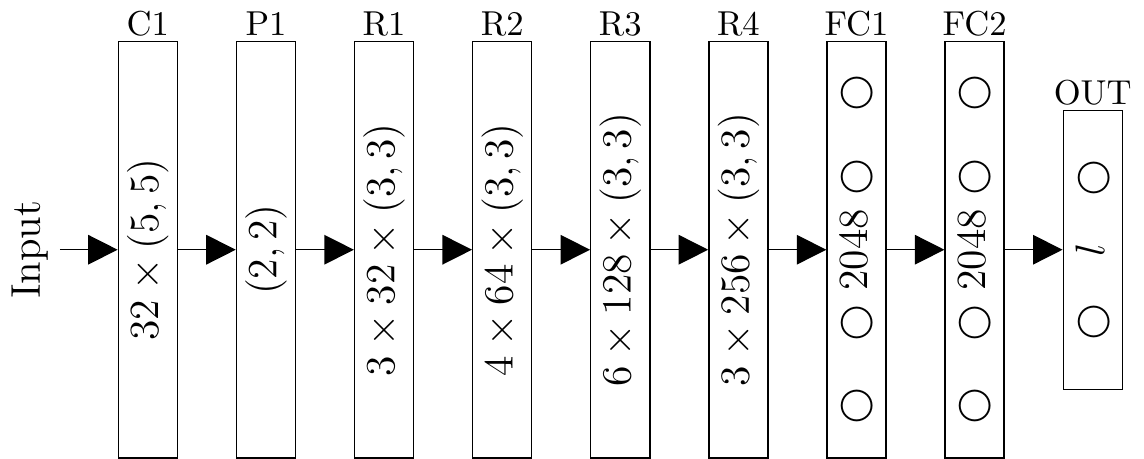}
\caption{Residual network architecture used for ParamNet in Sec.~\ref{sec:eval_3}}
\label{fig:resnet}
\end{subfigure}
\caption{\small Network architectures. {\bf C:} convolutional layer with feature maps indicated and kernel size in brackets, {\bf FC:} fully-connected layer with number of units indicated, {\bf OUT:} linear output layer with $l$ output units (matching the dimensionality of variable to be regressed), {\bf P:} max pooling with kernel size in brackets,  {\bf R:} residual module with the number of bottleneck blocks, the number of feature maps (bottleneck layer), and kernel size of bottleneck layers indicated.}
\label{fig:networks}
\vspace{-0.6cm}
\end{figure}

\subsection{Kinematic Hand Model Layer for Arbitrary Hand Shapes}
\label{sec:KinLayer}

The kinematic hand model layer implements the forward kinematics of the hand and therefore represents a mapping from hand parameters $\boldsymbol{\Lambda}$ to 3D joint locations $\tilde{\boldsymbol{J}}$. It is parameter free. In combination with the ParamNet, the resulting PoseNet is trained end-to-end. 

\begin{figure}[!h]
\centering
\begin{subfigure}[b]{0.19\textwidth}
\centering
\begin{tikzpicture}

\draw [arrows={-Triangle[length=0.2cm, width=0.2cm]}] (0.5*2/3,4.5*2/3) -- (1.5*2/3,4.5*2/3);
\draw [arrows={-Triangle[length=0.2cm, width=0.2cm]}] (0.5*2/3,4.5*2/3) -- (0.5*2/3,5.5*2/3);
\draw [arrows={-Triangle[length=0.2cm, width=0.2cm]}] (0.5*2/3,4.5*2/3) -- (-0.5*2*0.707/3,4.5*2/3-0.707*2/3);
\draw (1*2/3,5*2/3) node[thick] {$\boldsymbol{b}$};
\draw[black,thick,fill=white] (-0.125*2/3,5.625*2/3) circle (1.7ex);
\draw (-0.125*2/3,5.625*2/3) node[thick] {6D};

\draw[gray,thick,fill=gray] (0,0) circle (0.75ex);
\draw [arrows={-Triangle[length=0.2cm, width=0.2cm]},red] (0.5*2/3,4.5*2/3) -- (0,0);
\draw (-0.25*2/3,2.25*2/3) node[thick] {{\color{red}{}$\boldsymbol{v}_{\text{W}}$}};

\draw[blue,thick,fill=blue] (1.5*2/3,0.5*2/3) circle (0.75ex);
\draw [arrows={-Triangle[length=0.2cm, width=0.2cm]},red] (0.5*2/3,4.5*2/3) -- (1.5*2/3,0.5*2/3);
\draw (1.375*2/3,2.5*2/3) node[thick] {{\color{red}{}$\boldsymbol{v}_{\text{T}}$}};

\draw[blue,thick,fill=blue] (1.5*2/3,4*2/3) circle (0.75ex);
\draw [arrows={-Triangle[length=0.2cm, width=0.2cm]},red] (0.5*2/3,4.5*2/3) -- (1.5*2/3,4*2/3);
\draw (1.125*2/3,3.75*2/3) node[thick] {{\color{red}{}$\boldsymbol{v}_{\text{I}}$}};

\draw[blue,thick,fill=blue] (0.5*2/3,4.5*2/3) circle (0.75ex);

\draw[blue,thick,fill=blue] (-0.5*2/3,4.25*2/3) circle (0.75ex);
\draw [arrows={-Triangle[length=0.2cm, width=0.2cm]},red] (0.5*2/3,4.5*2/3) -- (-0.5*2/3,4.25*2/3);
\draw (-0.25*2/3,4.75*2/3) node[thick] {{\color{red}{}$\boldsymbol{v}_{\text{R}}$}};

\draw[blue,thick,fill=blue] (-1.5*2/3,3.5*2/3) circle (0.75ex);
\draw [arrows={-Triangle[length=0.2cm, width=0.2cm]},red] (0.5*2/3,4.5*2/3) -- (-1.5*2/3,3.5*2/3);
\draw (-1.25*2/3,4*2/3) node[thick] {{\color{red}{}$\boldsymbol{v}_{\text{P}}$}};

\end{tikzpicture}
\caption{Hand and finger bases}
\label{fig:handbase}
\end{subfigure}
\begin{subfigure}[b]{0.28\textwidth}
\centering
\begin{tikzpicture}

\draw [line width=0.5mm] (0,-0.5) -- (0,0);

\draw [line width=0.5mm] (0,0) -- (0.75,0.25);
\draw [red,line width=0.5mm] (0.75,0.25) -- (2,0.75);
\draw [red,line width=0.5mm] (2,0.75) -- (2.75,1.375);
\draw [red,line width=0.5mm] (2.75,1.375) -- (3,2);

\draw [line width=0.5mm] (0,0) -- (0.75,2);
\draw [red,line width=0.5mm] (0.75,2) -- (1.125,3);
\draw [red,line width=0.5mm] (1.125,3) -- (1.38,3.675);
\draw [red,line width=0.5mm] (1.38,3.675) -- (1.585,4.225);

\draw [line width=0.5mm] (0,0) -- (0.25,2.125);
\draw [red,line width=0.5mm] (0.25,2.125) -- (0.25,3.25);
\draw [red,line width=0.5mm] (0.25,3.25) -- (0.25,4);
\draw [red,line width=0.5mm] (0.25,4) -- (0.25,4.55);

\draw [line width=0.5mm] (0,0) -- (-0.25,2);
\draw [red,line width=0.5mm] (-0.25,2) -- (-0.375,3);
\draw [red,line width=0.5mm] (-0.375,3) -- (-0.46,3.675);
\draw [red,line width=0.5mm] (-0.46,3.675) -- (-0.53,4.225);

\draw [line width=0.5mm] (0,0) -- (-0.75,1.75);
\draw [red,line width=0.5mm] (-0.75,1.75) -- (-1.075,2.5);
\draw [red,line width=0.5mm] (-1.075,2.5) -- (-1.29,3);
\draw [red,line width=0.5mm] (-1.29,3) -- (-1.475,3.425);

\draw[gray,thick,fill=white] (0,0) circle (1.2ex);
\draw (0,0) node[thick] {{\color{gray} 0}};

\draw[blue,thick,fill=white] (0.75,0.25) circle (1.2ex);
\draw (0.75,0.25) node[thick] {{\color{blue} 2}};
\draw[violet,thick,fill=white] (2,0.75) circle (1.2ex);
\draw (2,0.75) node[thick] {{\color{violet} 2}};
\draw[orange,thick,fill=white] (2.75,1.375) circle (1.2ex);
\draw (2.75,1.375) node[thick] {{\color{orange} 1}};
\draw[cyan,thick,fill=white] (3,2) circle (1.2ex);
\draw (3,2) node[thick] {{\color{cyan} 0}};
\draw (1,-0.125) node[color=blue] {$\theta_{\text{T},1}$, $\theta_{\text{T},2}$};
\draw (2.25,0.375) node[color=violet] {$\theta_{\text{T},3}$, $\theta_{\text{T},4}$};
\draw (2.75,0.95) node[color=orange] {$\theta_{\text{T},5}$};
\draw (1.25,0.75) node[color=red] {$r_{\text{T},2}$};
\draw (2.125,1.25) node[color=red] {$r_{\text{T},4}$};
\draw (2.5,1.75) node[color=red] {$r_{\text{T},5}$};

\draw[blue,thick,fill=white] (0.75,2) circle (1.2ex);
\draw (0.75,2) node[thick] {{\color{blue} 3}};
\draw[violet,thick,fill=white] (1.125,3) circle (1.2ex);
\draw (1.125,3) node[thick] {{\color{violet} 1}};
\draw[orange,thick,fill=white] (1.38,3.675) circle (1.2ex);
\draw (1.38,3.675) node[thick] {{\color{orange} 1}};
\draw[cyan,thick,fill=white] (1.585,4.225) circle (1.2ex);
\draw (1.585,4.225) node[thick] {{\color{cyan} 0}};

\draw[blue,thick,fill=white] (0.25,2.125) circle (1.2ex);
\draw (0.25,2.125) node[thick] {{\color{blue} 3}};
\draw[violet,thick,fill=white] (0.25,3.25) circle (1.2ex);
\draw (0.25,3.25) node[thick] {{\color{violet} 1}};
\draw[orange,thick,fill=white] (0.25,4) circle (1.2ex);
\draw (0.25,4) node[thick] {{\color{orange} 1}};
\draw[cyan,thick,fill=white] (0.25,4.55) circle (1.2ex);
\draw (0.25,4.55) node[thick] {{\color{cyan} 0}};

\draw[blue,thick,fill=white] (-0.25,2) circle (1.2ex);
\draw (-0.25,2) node[thick] {{\color{blue} 3}};
\draw[violet,thick,fill=white] (-0.375,3) circle (1.2ex);
\draw (-0.375,3) node[thick] {{\color{violet} 1}};
\draw[orange,thick,fill=white] (-0.46,3.675) circle (1.2ex);
\draw (-0.46,3.675) node[thick] {{\color{orange} 1}};
\draw[cyan,thick,fill=white] (-0.53,4.225) circle (1.2ex);
\draw (-0.53,4.225) node[thick] {{\color{cyan} 0}};

\draw[blue,thick,fill=white] (-0.75,1.75) circle (1.2ex);
\draw (-0.75,1.75) node[thick] {{\color{blue} 3}};
\draw[violet,thick,fill=white] (-1.075,2.5) circle (1.2ex);
\draw (-1.075,2.5) node[thick] {{\color{violet} 1}};
\draw[orange,thick,fill=white] (-1.29,3) circle (1.2ex);
\draw (-1.29,3) node[thick] {{\color{orange} 1}};
\draw[cyan,thick,fill=white] (-1.475,3.425) circle (1.2ex);
\draw (-1.475,3.425) node[thick] {{\color{cyan} 0}};

\draw (2.75,4.25) node[thick] {{\color{cyan} TIP}};
\draw (2.75,3.8) node[thick] {{\color{orange} DIP}};
\draw (2.75,3.35) node[thick] {{\color{violet} PIP}};
\draw (2.75,2.95) node[thick] {{\color{blue} MCP}};
\draw (2.75,2.5) node[thick] {{\color{gray} WRIST}};

\end{tikzpicture}
\caption{DH finger modeling}
\label{fig:handdh}
\end{subfigure}
\caption{\small Kinematic hand modeling: {\bf (a)} 6D hand base $\boldsymbol{b}$ and finger/wrist vectors $\boldsymbol{v}_i$, $i\in \lbrace\text{T},\text{I},\text{R},\text{P}\rbrace$/$\boldsymbol{v}_\text{W}$ (red) {\bf (b)} Number of DoF associated with the individual joints indicated in the circles. Finger angles $\theta_{i,n}$ and bone lengths $r_{i,n}$ exemplary indicated for the thumb.}
\label{fig:hand}
\vspace{-0.7cm}
\end{figure}
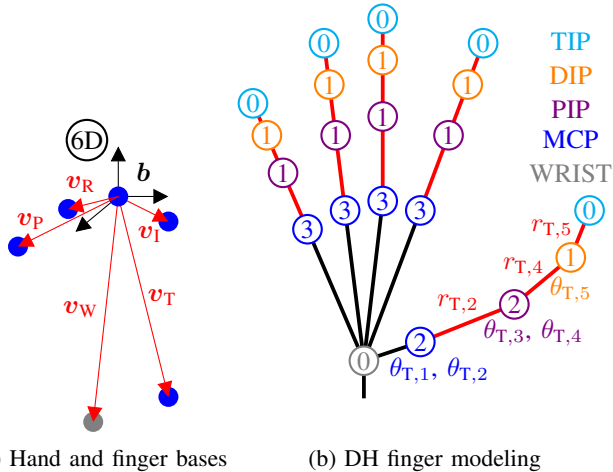

The inputs $\boldsymbol{\Lambda}$ to the kinematic layer divide into four groups (see Fig.~\ref{fig:hand}) listed below. The index $n \in \lbrace 1,2,3,4,5 \rbrace$ indicates the associated Denavit-Hartenberg (DH) transformation of the respective finger (counted from MCP to TIP joint).
\begin{itemize}
\item 6D global pose and orientation of the middle finger MCP joint, which we define as the hand base $\boldsymbol{b}$ (6D)
\item Four 3D vectors $\boldsymbol{v}_i$ ($i \neq \text{M}$) from the hand base $\boldsymbol{b}$ to the remaining finger bases and one 3D vector $\boldsymbol{v}_\text{W}$ from the hand base to the wrist base (15D)
\item 15 finger bone lengths $r_{i,n}$, (red in Fig.~\ref{fig:handdh}) (15D)
\item 25 finger joint angles $\theta_{i,n}$ (25D)
\end{itemize} 

The ParamNet in front of the kinematic layer regresses the 61 hand parameters $\boldsymbol{\Lambda}$. The 3D joint locations $\tilde{\boldsymbol{J}}$ are calculated using the hand parameters $\boldsymbol{\Lambda}$ by chaining the appropriate transformation matrices. Each joint is considered to be the origin of its own local coordinate system. In order to back-transform these local coordinates to global world coordinates, appropriate kinematic transformations must be applied. For a 3D joint location $\tilde{\boldsymbol{j}}_{i,k}$ this transformation is denoted as
\begin{equation}
\tilde{\boldsymbol{j}}_{i,k} = \boldsymbol{T}_{\text{BASE}}\left(\boldsymbol{b}\right)\boldsymbol{T}_{\text{VEC},i}\left(\boldsymbol{v}_i\right)\prod_{n=1}^{N_{\text{DH},i,k}}\boldsymbol{T}_{\text{DH},n}\left(\theta_{i,n},r_{i,n}\right)\begin{pmatrix}
0\\ 0\\ 0\\ 1
\end{pmatrix}.
\end{equation}
The kinematics of the fingers are modeled using the DH convention with $\theta$ being the finger joint angles $\theta_{i,n}$ and $r$ being the finger bone lengths $r_{i,n}$. The $\alpha$ and $d$ parameters are fixed in such a way that valid hand kinematics result. Therefore, in the case $k\in \lbrace \text{PIP},\text{DIP},\text{TIP} \rbrace$, first, $N_{\text{DH},i,k}$ DH transformations $\boldsymbol{T}_{\text{DH},n}\left(\theta_{i,n},r_{i,n}\right)$ are applied to the local joint coordinate $\left( 0,0,0,1\right)^T$. $N_{\text{DH},i,k}$ is the number of DoF along the kinematic chain from the joint $\tilde{\boldsymbol{j}}_{i,k}$ to the MCP joint of the corresponding finger (DoF per joint shown in Fig.~\ref{fig:handdh}). In this way, the PIP, DIP, and TIP joints of all fingers are transformed into the local coordinate system of the MCP joint of the respective finger.

The resulting coordinates of all finger joints (except middle finger joints) as well as the wrist joint coordinates are then transformed into the local coordinate system of the middle finger MCP joint (the hand base). To do so, the coordinates are translated by the finger vectors $\boldsymbol{v}_i$ and the wrist vector $\boldsymbol{v}_\text{W}$, respectively, using the transformation matrix $\boldsymbol{T}_{\text{VEC},i}\left(\boldsymbol{v}_i\right)$. Finally, the joint coordinates are transformed into world coordinates by applying the homogeneous transformation matrix $\boldsymbol{T}_{\text{BASE}}\left(\boldsymbol{b}\right)$, which is parametrized with the 6D hand base translation and rotation parameters $\boldsymbol{b}$.

The procedure of chaining transformation matrices is computationally efficient, as many intermediate results can be re-used for the coordinate transformation of other joints along the same finger. We do not have to explicitly specify the gradients because we implement our approach in TensorFlow \cite{Abadi2016}, which supports auto-differentiation.

In order to force the finger joint angles $\theta_{i,n}$ to stay within physically valid limits, we adopt the idea of physical constraint losses on the estimated joint angles $\boldsymbol{\theta}^{\text{est}}$ from \cite{Zhou2016}. The physical constraint loss term which penalizes violations of upper and lower joint angle limits $\boldsymbol{\theta}_{\text{up}}$ and $\boldsymbol{\theta}_{\text{low}}$ is
\begin{equation}
\begin{split}
L_{\text{constr}} = \sum_{\text{batch}}(\vert\vert\min\left(\left(\boldsymbol{\theta}^{\text{est}}-\boldsymbol{\theta}_{\text{low}}\right),0\right)\vert\vert^2\\ 
+\vert\vert\max\left(\left(\boldsymbol{\theta}^{\text{est}}-\boldsymbol{\theta}_{\text{up}}\right),0\right)\vert\vert^2)
\end{split}.
\end{equation}
It is added to the standard squared euclidean joint position loss of the PoseNet:
\begin{equation}
L_{\text{joint}} = \frac{1}{2}\sum_{\text{batch}}\sum_{\text{joints}}\vert\vert \boldsymbol{j}_{i,k}^{\text{est}}-\boldsymbol{j}_{i,k}^{\text{gt}}\vert\vert^2
\end{equation}

\section{EVALUATION}

Our different hand pose estimation components are evaluated in various combinations on the Hands 2017 challenge dataset \cite{Yuan2017b} and the NYU \cite{Tompson2014} dataset. The Hands 2017 challenge dataset, which is composed from parts of the Big Hand 2.2M dataset \cite{Yuan2017a} and the First-person Hand Action Dataset (FHAD) \cite{Garcia-Hernando2017}, is currently the largest and most diverse dataset available. Its training set contains 957032 depth images of five different hands. Therefore, it is suited for learning to regress hand parameters for various hand shapes. The test set consists of 295510 depth images of ten different hand shapes, of which five are the same as in the training set and five are entirely new. The NYU dataset contains 72757 training images of a single subject's hand and 8252 test images that include a second hand shape besides the one from the training set.

Our approach is implemented using TensorFlow \cite{Abadi2016}. The networks are trained on a PC with an AMD FX-4300/Intel Core i7-860 CPU and an nVidia GeForce GTX1060 6GB GPU. For network training, the Adam optimizer is used with a learning rate of \num{1e-4}. The batch size is 32.

\subsection{Kinematic Hand Model Layer for Arbitrary Hand Shapes}
\label{sec:eval_1}

First, the effectiveness of the kinematic hand model layer implementation with variable bone lengths and finger base positions is tested standalone. Therefore, three different models are trained for 50 epochs on the Hands 2017 challenge training dataset: 1) a basic CNN as ParamNet combined with our new variable hand shape kinematic layer (\textit{Variable Hand CNN}), 2) a basic CNN with the kinematic layer with bone lengths and finger base positions fixed to the mean values across the training set (\textit{Fixed Hand CNN}), and 3) a basic CNN without any kinematic layer that directly predicts the 21 3D joint locations (\textit{Direct CNN}) as a baseline.

The evaluation results on the Hands 2017 challenge test set are listed in Tab.~\ref{tab:handmodel} and Tab.~\ref{tab:constvio}. In Tab.~\ref{tab:handmodel}, three different average per joint 3D location errors
\begin{equation}
e_{\text{joint}} = \frac{1}{N_{\text{images}}}\frac{1}{N_{\text{joints}}}\sum_{\text{images}}\sum_{\text{joints}}\vert\vert \boldsymbol{j}_{i,k}^{\text{est}}-\boldsymbol{j}_{i,k}^{\text{gt}}\vert\vert
\end{equation}
are given: 1) the average across the complete test set (\textit{Avg test}), 2) the average across the test set images of hand shapes seen during training (\textit{Seen test}), and 3) the average across the test set images of unseen hand shapes (\textit{Unseen test}). In Tab.~\ref{tab:constvio}, the frequency and severity of joint angle limit violations are evaluated. For the \textit{Variable Hand CNN}, the joint angles are available as outputs of the CNN. In the case of the \textit{Direct CNN}, the joint angles are calculated from the estimated joint locations using inverse kinematics.

As seen in Tab.~\ref{tab:handmodel}, making the bone lengths and finger base positions of the kinematic hand model layer variables that are regressed by the ParamNet CNN decreases the average test set error (\textit{Avg test}) by \SI{2.33}{\milli\metre}. The hand pose estimation approach is now able to generalize towards arbitrary hand shapes. However, the estimation accuracy is \SI{0.44}{\milli\metre} lower than the one of the baseline CNN without any kinematic hand model layer.
\begin{table}[htb]
\begin{center}
\vspace{-0.3cm}
\caption{Comparison of hand model layer implementations}
\begin{tabular}{ l | l | l | l }
Approach & Avg test & Seen test & Unseen test\\ \hline
Fixed Hand CNN & \SI{20.39}{\milli\metre} & \SI{16.90}{\milli\metre} & \SI{23.29}{\milli\metre}\\
Variable Hand CNN & \SI{18.06}{\milli\metre} & \SI{14.89}{\milli\metre} & \SI{20.70}{\milli\metre}\\
Direct CNN & \SI{17.62}{\milli\metre} & \SI{14.57}{\milli\metre} & \SI{20.16}{\milli\metre}\\
\end{tabular}
\label{tab:handmodel}
\end{center}
\vspace{-0.5cm}
\end{table}

Only considering the joint location error $e_{\text{joint}}$, the usage of a kinematic hand model layer is not recommendable. However, the \textit{Variable Hand CNN} with physical constraint losses $L_{\text{constr}}$ on the estimated joint angles trades pure accuracy for a higher physical validity of the estimated hand pose. Table~\ref{tab:constvio} shows that the \textit{Variable Hand CNN} reduces the percentage of joint limit violations to almost zero, compared to \SI{6.35}{\percent} in the case of the \textit{Direct CNN}. Furthermore, the average deviation between the joint angle limit and the actual joint angle, in case of a violation, reduces from \SI{44.09}{\degree} to \SI{2.34}{\degree}.
\begin{table}[htb]
\begin{center}
\caption{Joint angle constraint violation frequency and severity with and without kinematic hand model layer}
\begin{tabular}{ l | l | l | l }
Approach & \pbox{20cm}{Violated \\ joint limits} & \pbox{20cm}{Avg violation in\\ case of violation} & \pbox{20cm}{Avg violation\\ in total}\\ \hline
Direct CNN & \SI{6.350}{\percent} & \SI{44.09}{\degree} & \SI{2.80}{\degree}\\
Variable Hand CNN & \SI{0.025}{\percent} & \SI{2.34}{\degree} & \SI{0.00}{\degree}\\
\end{tabular}
\label{tab:constvio}
\end{center}
\vspace{-0.8cm}
\end{table}

\subsection{Appearance Normalization Pipeline}
\label{sec:eval_2}

This section evaluates the performance of our proposed appearance normalization pipeline. First, it is shown that Box-, Rot-, and ScaleNet are able to estimate the transformation parameters $\boldsymbol{t}$, $\alpha_z$, and $s$. Therefore, Box-, Rot-, and ScaleNet are trained standalone as well as in combination for 50 epochs on the Hands 2017 challenge dataset. 

The average errors in the estimated translation $\vert\vert\boldsymbol{t}^{\text{est}}-\boldsymbol{t}^{\text{gt}}\vert\vert$, rotation $\vert\alpha_z^\text{est}-\alpha_z^{\text{gt}}\vert$ (taking into account the angle wrap around), and scaling $\vert s^{\text{est}}-s^{\text{gt}}\vert$ are evaluated across the 7040 image validation set, which we separate from the Hands 2017 challenge training dataset. Table~\ref{tab:varred} lists these \textit{validation errors} that represent the remaining transformation error after normalization and compares them to the \textit{average ground truth} transformation values $\vert\vert\boldsymbol{t}^{\text{gt}}\vert\vert$ , $\vert\alpha_z^\text{gt}\vert$, and $\vert 1-s^{\text{gt}} \vert$ across the Hands 2017 challenge training set before normalization. 

The results show that Box-, Rot-, and ScaleNet can be used to normalize images relatively well, even under large amount of transformation. Furthermore, the results show that cascading the individual networks is beneficial. Training RotNet on top of BoxNet, reduces the error in $\alpha_z$ from \SI{15.33}{\degree} to \SI{14.15}{\degree}. A ScaleNet trained on top of those two networks lowers the error in $s$ from \num{0.0201} to \num{0.0144}.
\begin{table}[htb]
\begin{center}
\vspace{-0.1cm}
\caption{Estimation errors in translation, rotation, and scaling of Box-, Rot-, and ScaleNet compared to ground truth}
\begin{tabular}{ l | l | l }
Network(s) & Average ground truth value & Validation error\\ \hline
BoxNet & \SI{56.43}{\milli\metre} & \SI{4.95}{\milli\metre}\\
RotNet & \SI{82.99}{\degree} & \SI{15.33}{\degree}\\
Box+RotNet & \SI{82.99}{\degree} & \SI{14.15}{\degree}\\
ScaleNet & \num{0.0492} & \num{0.0201}\\
Box+Rot+ScaleNet & \num{0.0492} & \num{0.0144}\\
\end{tabular}
\label{tab:varred}
\end{center}
\vspace{-0.6cm}
\end{table}

Although the errors on the actual test set are unknown \footnote{The Hands 2017 challenge test set comes without ground truth annotation. Only joint location estimates can be evaluated online.}, Box-, Rot-, and ScaleNet still reduce the variance in the appearance of the hand images. This can be seen from the results listed in Tab.~\ref{tab:cascade}.
\begin{table}[htb]
\vspace{-0.2cm}
\begin{center}
\caption{Normalization pipeline performance study}
\begin{tabular}{ l | l | l | l }
Approach & Avg test & Seen test & Unseen test\\ \hline
Direct CNN & \SI{17.62}{\milli\metre} & \SI{14.57}{\milli\metre} & \SI{20.16}{\milli\metre}\\
... with BoxNet & \SI{16.61}{\milli\metre} & \SI{13.17}{\milli\metre} & \SI{19.48}{\milli\metre}\\
... with RotNet & \SI{16.22}{\milli\metre} & \SI{12.74}{\milli\metre} & \SI{19.13}{\milli\metre}\\
... with ScaleNet & \SI{17.29}{\milli\metre} & \SI{14.13}{\milli\metre} & \SI{19.92}{\milli\metre}\\
... with Box+RotNet & \SI{15.47}{\milli\metre} & \SI{11.82}{\milli\metre} & \SI{18.52}{\milli\metre}\\
... with Box+Rot+ScaleNet & \SI{15.40}{\milli\metre} & \SI{11.74}{\milli\metre} & \SI{18.46}{\milli\metre}\\ \hline
Variable Hand CNN & \SI{18.06}{\milli\metre} & \SI{14.89}{\milli\metre} & \SI{20.70}{\milli\metre}\\
... with Box+Rot+ScaleNet & \SI{16.64}{\milli\metre} & \SI{12.80}{\milli\metre} & \SI{19.48}{\milli\metre}\\
\end{tabular}
\label{tab:cascade}
\end{center}
\vspace{-0.5cm}
\end{table}

A \textit{Direct CNN} that directly regresses the 3D joint locations is trained for 50 epochs on top of Box-, Rot-, and ScaleNet, respectively. In each case, the average per joint estimation error on the test set decreases compared to the test error of the \textit{Direct CNN} standalone. The largest reduction is achieved by the RotNet $\left(\SI{1.4}{\milli\metre}\right)$, followed by the BoxNet $\left(\SI{1.01}{\milli\metre}\right)$, whereas the ScaleNet achieves a reduction of $\SI{0.33}{\milli\metre}$. Furthermore, the results of training the \textit{Direct CNN} on top of Box+RotNet and the complete pipeline (Box+Rot+ScaleNet) are given. The complete pipeline reduces the test error by $\SI{2.22}{\milli\metre}$ compared to the \textit{Direct CNN} standalone. The error reduction is larger for the seen hand shapes $\left(\SI{2.83}{\milli\metre}\right)$ than for the unseen $\left(\SI{1.7}{\milli\metre}\right)$. When combining the complete pipeline with the \textit{Variable Hand CNN}, the test set error decreases by $\SI{1.42}{\milli\metre}$ with a larger decrease for seen hand shapes $\left(\SI{2.09}{\milli\metre}\right)$ than for unseen $\left(\SI{1.22}{\milli\metre}\right)$.

\begin{figure}
\centering
\includegraphics[width=0.46\textwidth]{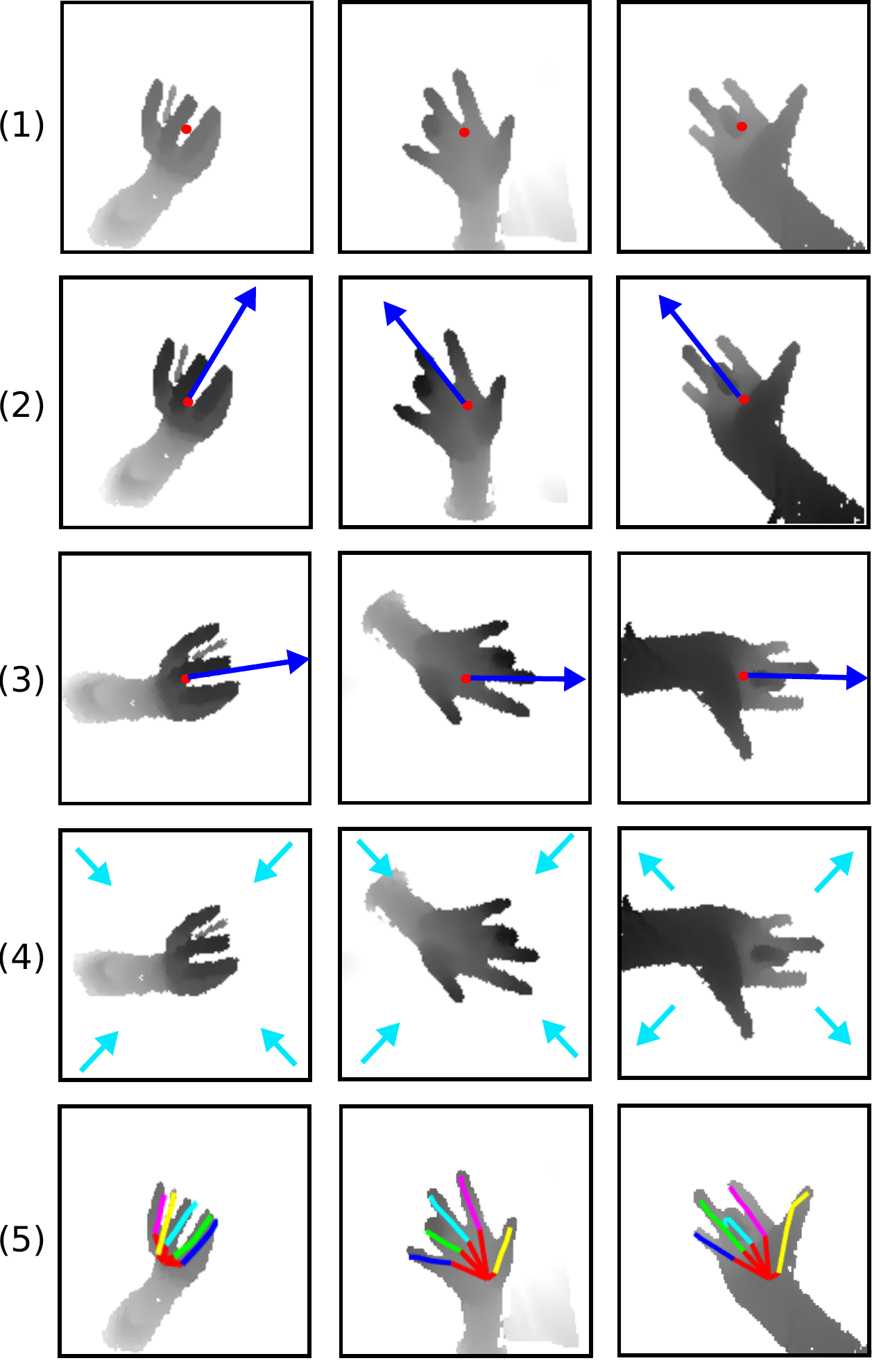}
\caption{\small Visualization of the individual appearance normalization pipeline stages for three different examples (columns). From top to bottom: (1) $128\times 128$ center crops of pre-processed and augmented input, red dot indicates image center. (2) Re-centered images after BoxNet, blue arrow indicates middle finger direction. (3) Rotated images after RotNet. (4) Re-scaled images after ScaleNet (visual effect not too obvious). (5) The estimated hand poses, which are back-transformed to the original coordinate system of (1).}
\label{fig:visu}
\vspace{-0.6cm}
\end{figure}

Qualitative results that visualize the individual stages of the appearance normalization pipeline are shown in Fig.~\ref{fig:visu}.

\subsection{Residual Network for Higher Accuracy}
\label{sec:eval_3}

For a higher overall accuracy, the CNN used as ParamNet, to regress the hand model parameters or directly the 3D joint locations, is replaced with the more powerful ResNet architecture (Fig.~\ref{fig:resnet}). The results of training the ResNet architecture with the kinematic hand model layer for variable hand shape standalone (\textit{Variable Hand ResNet}) and on top of the appearance normalization pipeline for 50 epochs each on the Hands 2017 challenge dataset are shown in Tab.~\ref{tab:cascaderes}.
\begin{table}[htb]
\begin{center}
\caption{Comparison with other challenge submissions}
\begin{tabular}{ l | l | l | l }
Approach & Avg test & Seen test & Unseen test\\ \hline
Variable Hand ResNet & \SI{12.13}{\milli\metre} & \SI{9.95}{\milli\metre} & \SI{13.95}{\milli\metre}\\
... with Box+Rot+ScaleNet & \SI{12.86}{\milli\metre} & \SI{9.95}{\milli\metre} & \SI{15.29}{\milli\metre}\\ \hline
mks0601 \cite{Moon2017} & \SI{9.95}{\milli\metre} & \SI{6.97}{\milli\metre} & \SI{12.43}{\milli\metre}\\
slivorezzz & \SI{9.97}{\milli\metre} & \SI{7.55}{\milli\metre} & \SI{12.00}{\milli\metre}\\
oasis & \SI{11.30}{\milli\metre} & \SI{8.86}{\milli\metre} & \SI{13.33}{\milli\metre}\\
THU\_EE\_VCLab & \SI{11.70}{\milli\metre} & \SI{9.15}{\milli\metre} & \SI{13.83}{\milli\metre}\\
Vanora & \SI{11.91}{\milli\metre} & \SI{9.55}{\milli\metre} & \SI{13.89}{\milli\metre}\\
NAIST\_RV & \SI{13.18}{\milli\metre} & \SI{10.64}{\milli\metre} & \SI{15.30}{\milli\metre}\\
strawberryfg & \SI{13.91}{\milli\metre} & \SI{9.87}{\milli\metre} & \SI{17.27}{\milli\metre}\\
rvhand & \SI{14.57}{\milli\metre} & \SI{12.07}{\milli\metre} & \SI{16.65}{\milli\metre}\\
mmadadi & \SI{14.74}{\milli\metre} & \SI{11.87}{\milli\metre} & \SI{17.14}{\milli\metre}\\
maxo & \SI{16.88}{\milli\metre} & \SI{13.27}{\milli\metre} & \SI{19.90}{\milli\metre}\\
Franziska & \SI{18.57}{\milli\metre} & \SI{15.39}{\milli\metre} & \SI{21.23}{\milli\metre}\\
Baseline & \SI{19.71}{\milli\metre} & \SI{14.58}{\milli\metre} & \SI{23.98}{\milli\metre}\\
\end{tabular}
\label{tab:cascaderes}
\end{center}
\vspace{-0.8cm}
\end{table}

The \textit{Variable Hand ResNet} reduces the average test error by \SI{5.93}{\milli\metre} compared to the \textit{Variable Hand CNN}. Furthermore, our approach is compared to other challenge submissions.\footnote{The test errors shown in Tab.~\ref{tab:cascaderes} are taken from the online leader-board on February 4 2018. \url{https://competitions.codalab.org/competitions/17356\#results}}
Our approach performs well, only being two to three millimeters worse in accuracy than the leading approaches. Compared to the best performing approach of \textit{mks0601} \cite{Moon2017}, our approach is conceptually and especially computationally much simpler. While the approach presented in \cite{Moon2017} carries out inference with \num{3.5} frames per second on a single GPU, our approach achieves \num{838} frames per second and therefore has real-time capability. Apart from that, our approach offers specific benefits: First, the kinematic layer offers access to the hand parameters, which might be useful for hand tracking applications. Second, physically valid hand pose estimates can be enforced by physical constraint loss terms on the respective hand parameters.

Adding the appearance normalization pipeline is, however, not beneficial in combination with the ResNet. The average test error is \SI{0.73}{\milli\metre} higher. While the error for the seen hand shapes is the same, the error for the unseen hand shapes is \SI{1.34}{\milli\metre} higher. The generalization ability is decreased. Possible reasons for this behavior are discussed in Sec.~\ref{sec:discussion}.

\subsection{Results on the NYU Dataset}
\label{sec:NYU}

The \textit{Variable Hand CNN} as well as the \textit{Variable Hand ResNet} with and without the appearance normalization pipeline are, furthermore, trained on the NYU dataset for 100 epochs each. We use all three views (frontal + side views) of the images in the NYU training dataset for training. Although the NYU dataset annotates 36 joints, we use the same 14 joints for evaluation as most earlier works like \cite{Tompson2014} or \cite{Moon2017}.

Table~\ref{tab:nyu} lists the average per joint 3D location error $e_{\text{joint}}$ across the test set (\textit{Test error}). Apart from our own results, the estimation error values of several other prominent approaches are given. Although the NYU training set only contains a single hand shape, our \textit{Variable Hand CNN} approach outperforms the kinematic hand model based approach of Zhou et al. \cite{Zhou2016} by \SI{0.6}{\milli\metre}. Using our ResNet architecture as ParamNet, the \textit{Variable Hand ResNet} reduces the average joint location error by another \SI{4.9}{\milli\metre}.
\begin{table}[htb]
\begin{center}
\caption{3D joint location errors on NYU}
\begin{tabular}{ l | l }
Approach & Test error\\ \hline
{\bf Variable Hand CNN} & {\bf \SI{16.3}{\milli\metre}}\\
... with Box+Rot+ScaleNet & \SI{13.0}{\milli\metre}\\ 
Variable Hand ResNet & \SI{11.4}{\milli\metre}\\
... with Box+Rot+ScaleNet & \SI{11.0}{\milli\metre}\\ \hline
Crossing Nets \cite{Wan2017a} & \SI{15.5}{\milli\metre}\\
{\bf  DeepModel} \cite{Zhou2016} & {\bf  \SI{16.9}{\milli\metre}}\\
DeepPrior \cite{Oberweger2015a} & \SI{19.8}{\milli\metre}\\
DeepPrior++ \cite{Oberweger2017} & \SI{12.3}{\milli\metre}\\
Feedback \cite{Oberweger2015b} & \SI{16.2}{\milli\metre}\\
Global to Local \cite{Madadi2017a} & \SI{15.6}{\milli\metre}\\
Hand3D \cite{Deng2017} & \SI{17.6}{\milli\metre}\\
HMDN \cite{Ye2017} & \SI{16.3}{\milli\metre}\\
Pose-REN \cite{Chen2017b} & \SI{11.8}{\milli\metre}\\
REN \cite{Guo2017b} & \SI{12.7}{\milli\metre}\\
SGN \cite{Ye2017} & \SI{15.9}{\milli\metre}\\
V2V-PoseNet \cite{Moon2017} & \SI{8.4}{\milli\metre}\\
\end{tabular}
\label{tab:nyu}
\end{center}
\vspace{-0.8cm}
\end{table}

Finally, the \textit{Variable Hand CNN} and the \textit{Variable Hand ResNet} are trained on top of our appearance normalization pipeline. This time, we are able to provide test set errors for the transformation parameters. Table ~\ref{tab:varrednyu} shows the \textit{average ground truth} transformation values and the remaining transformation error after normalization evaluated across the test set. This shows that our appearance normalization networks also work reasonably on a proper test set. 
\begin{table}[!h]
\begin{center}
\caption{Transformation parameter estimation on NYU}
\begin{tabular}{ l | l | l }
Network(s) & Average ground truth value & Test error\\ \hline
BoxNet & \SI{40.40}{\milli\metre} & \SI{10.98}{\milli\metre}\\
Box+RotNet & \SI{72.72}{\degree} & \SI{18.54}{\degree}\\ 
Box+Rot+ScaleNet & \num{0.0807} & \num{0.0498}\\
\end{tabular}
\label{tab:varrednyu}
\end{center}
\vspace{-0.4cm}
\end{table}

Combining the appearance normalization pipeline with the \textit{Variable Hand CNN} reduces the average joint location error by \SI{3.3}{\milli\metre}. Contrary to the results on the Hands 2017 challenge dataset, the appearance normalization pipeline also decreases the estimation error of the \textit{Variable Hand ResNet} (by \SI{0.4}{\milli\metre}) and therefore shows its effectiveness in combination with a ResNet. We outperform most of the approaches on the NYU dataset. The \textit{V2V-PoseNet} \cite{Moon2017} is, as discussed in Sec.~\ref{sec:eval_3}, computationally heavier than our approach.

\subsection{Discussion: Appearance Normalization Pipeline}
\label{sec:discussion}

The appearance normalization shows a mixed performance when combined with a ResNet. While it is beneficial on the NYU dataset, it decreases the estimation accuracy on the Hands 2017 challenge dataset. There are three possible reasons for this phenomenon: 1) the appearance normalization pipeline takes out too much of the regularization provided by the translation, rotation, and scale augmentation of which the ResNet needs more, due to its higher model capacity, than the CNN. On the NYU dataset, this lack of regularization is not so dramatic, as the NYU dataset has less variation in hand pose and viewpoint. 2) a ResNet standalone can learn better normalization features than the hand-crafted ones provided by the pipeline. Again, this might be less of a problem on the NYU dataset, as it is "easier" and the pipeline might therefore perform relatively better. 3) the Hands 2017 challenge test set is constituted differently than the training set. It contains a higher fraction of egocentric views and some hand object interaction images that are not present in the training set at all. This hypothesis is supported by the fact that the joint location error $e_{\text{joint}}$ on the unseen training images of seen hand shapes we use for validation is \SI{1.06}{\milli\metre} lower with the appearance normalization pipeline than without. The error on the test images of seen hand shapes is identical with and without the pipeline. So, using unseen test images of seen hand shapes for evaluation results in a relatively worse behavior of the appearance normalization pipeline in combination with the ResNet than using unseen training images of seen hand shapes.

\section{CONCLUSION}

In this work, we increase the estimation accuracy of kinematic hand-model-based hand pose estimation by treating hand shape parameters as variables to be regressed by a neural network so that the approach can generalize to arbitrary hand shapes. In combination with a ResNet, we can achieve state-of-the-art estimation accuracy while maintaining real-time capability. Apart from that, we show that it is possible to learn transformation parameters though a cascade of neural networks in order to approximately normalize the appearance of hand images with respect to translation, rotation, and scaling. While increasing the estimation accuracy when combined with a shallow CNN, this appearance normalization pipeline shows a mixed performance when combined with deep residual networks. Therefore, a future work is to find ways to combine the appearance normalization pipeline in a more beneficial way with residual networks. 






\bibliographystyle{IEEEtran}
\bibliography{IEEEabrv,mybibfile}

\end{document}